\documentclass[11pt,a4paper]{article}
\usepackage{authblk}
\usepackage[hyperref]{emnlp2018}
\usepackage{times}
\usepackage{latexsym}

\usepackage[utf8]{inputenc} 
\usepackage[T1]{fontenc}    
\usepackage{url}            
\usepackage{booktabs}       
\usepackage{amsfonts}       
\usepackage{amssymb}
\usepackage{nicefrac}       
\usepackage{microtype}      
\usepackage{comment}
\usepackage{amsmath}
\usepackage{subcaption}
\usepackage{siunitx}

\usepackage{bbm}

\usepackage{float}

\usepackage{tikz}
\usetikzlibrary{positioning}
\usetikzlibrary{fit}

\aclfinalcopy 
\def\aclpaperid{2098} 

\title{Modeling Online Discourse with Coupled Distributed Topics}

\author[$\dagger$]{Nikita Srivatsan}
\author[$\ddagger$]{Zachary Wojtowicz}
\author[$\dagger$]{Taylor Berg-Kirkpatrick}
\affil[$\dagger$]{Language Technologies Institute\\ Carnegie Mellon University\\ \texttt{\{asrivats,tberg\}@cs.cmu.edu}}
\affil[$\ddagger$]{Social and Decision Sciences\\ Carnegie Mellon University\\ \texttt{zdw@andrew.cmu.edu}}

\date{}

\begin{document}

\maketitle

\begin{abstract}
In this paper, we propose a deep, globally normalized topic model that incorporates structural relationships connecting documents in socially generated corpora, such as online forums. Our model $(1)$ captures discursive interactions along observed reply links in addition to traditional topic information, and $(2)$ incorporates latent distributed representations arranged in a deep architecture, which enables a GPU-based mean-field inference procedure that scales efficiently to large data. We apply our model to a new social media dataset consisting of $13$M comments mined from the popular internet forum Reddit, a domain that poses significant challenges to models that do not account for relationships connecting user comments.  We evaluate against existing methods across multiple metrics including perplexity and metadata prediction, and qualitatively analyze the learned interaction patterns.
\end{abstract}

\section{Introduction}

\begin{figure*}[th!]
\begin{minipage}[c]{0.67\textwidth}
\includegraphics[width=\textwidth]{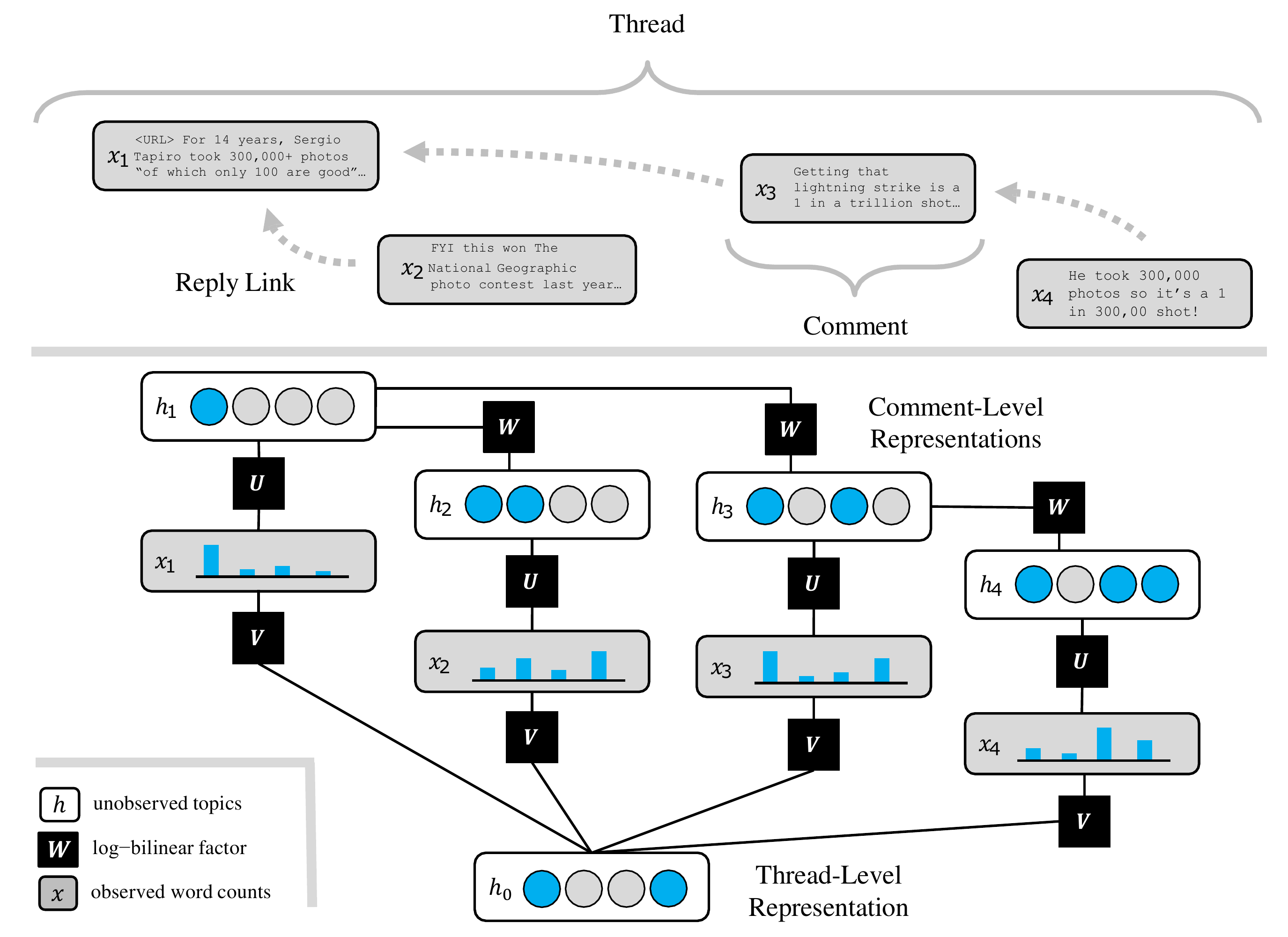}
\end{minipage} \hfill
\begin{minipage}[c]{0.3\textwidth}
\caption{DDTM factor graph for an example thread.
		 Each comment is modeled as an observed bag-of-words $x$ with topics represented
         by a latent binary vector $h$.
         Log-bilinear factors connect the latent and observed variables of each comment, and
         the latent variables of parent-child comment pairs along observed reply links. Biases
         are omitted for clarity.}
\label{fig:tabm}
\end{minipage}
\end{figure*}

Topic models have become one of the most common unsupervised methods for uncovering latent semantic information in natural language data, and have found a wide variety of applications across the sciences. However, many common models - such as Latent Dirichlet Allocation~\cite{LDA} - make an explicit exchangeability assumption that treats documents as independent samples from a generative prior, thereby ignoring important aspects of text corpora which are generated by non-ergodic, interconnected social systems. While the direct application of such models to datasets such as transcripts of The French Revolution~\cite{barron2017individuals} and discussions on Twitter~\cite{zhao2011comparing} have yielded sensible topics and exciting insights, their exclusion of document-to-document interactions imposes limitations on the scope of their applicability and the analyses they support. For instance, on many social media platforms, comments are short (the average Reddit comment is $10$ words long), making them difficult to treat as full documents, yet they do cohere as a collection, suggesting that contextual relationships should be considered. Moreover, analysis of social data is often principally concerned with understanding relationships between documents (such as question-asking and -answering), so a model able to capture such features is of direct scientific relevance.

To address these issues, we propose a design that models representations of comments jointly along observed reply links.
Specifically, we attach a vector of latent binary variables to each comment in a collection of social data, which in turn connect to each other according to the observed reply-link structure of the dataset. The inferred representations can provide information about the rhetorical moves and linguistic elements that characterize an evolving discourse. An added benefit is that while previous work such as Sequential LDA~\cite{du2012sequential} has focused on modeling a linear progression, the model we present applies to a more general class of acyclic graphs such as tree-structured comment threads ubiquitous on the web.

Online data can be massive, which presents a scalability issue for traditional methods.
Our approach uses latent binary variables similar to a Restricted Boltzmann Machine (RBM); related models such as Replicated Softmax (RS)~\cite{RS} have previously seen success in capturing latent properties of language, and found substantial speedups over previous methods due to their GPU amenable training procedure. RS was also shown to deal well with documents of significantly different length, another key characteristic of online data.
While RBMs permit exact inference, the additional coupling potentials present in our model make inference intractable. However, the choice of bilinear potentials and latent features admits a mean-field inference procedure which takes the form of a series of dense matrix multiplications followed by nonlinearities, which is particularly amenable to GPU computation and lets us scale efficiently to large data.

Our model outperforms LDA and RS baselines on perplexity and downstream tasks including metadata prediction and document retrieval when evaluated
on a new dataset mined from Reddit. We also qualitatively analyze the learned topics and discuss the social phenomena uncovered.

\section{Model}

We now present an overview of our model.
Specifically, it will take the probabilistic form
of an undirected graphical model whose architecture mirrors the tree structure of
the threads in our data.

\subsection{Motivating Dataset}

We evaluate on a corpus mined from Reddit, an internet forum which ranks as the fourth most trafficked site in the US~\cite{Alexa} and sees millions of daily comments~\cite{RedditBlog}. Discourse on Reddit follows a branching pattern, shown in Figure~\ref{fig:tabm}. The largest unit of discourse is a thread, beginning with a link to external content or a natural language prompt, posted to a relevant subreddit based on its subject matter. Users comment in response to the original post (OP), or to any other comment. The result is a structure which splits at many points into more specific or tangential discussions that while locally coherent may differ substantially from each other. The data reflect features of the underlying memory and network structure of the generating process; comments are serially correlated and highly cross-referential. We treat individual comments as ``documents'' under the standard topic modeling paradigm, but use observed reply structure to induce a tree of documents for every thread.

\subsection{Description of Discursive Distributed Topic Model}

We now introduce the Discursive Distributed Topic Model (DDTM) (illustrated in Figure \ref{fig:tabm}).
For each comment in the thread, DDTM assigns a latent vector of binary random variables (or \textit{bits}) that collectively form a distributed embedding of
the topical content of that comment; for instance, one bit might represent sarcastic language
while another might track usage of specific acronyms - a given comment could have any
combination of those features.
These representations are tied to those of parent and child comments via
coupling potentials (see Section~\ref{sec:prob}), which allow them to learn discursive properties by inducing a deep undirected network over the thread.
In order to encourage the model to use these \textit{comment-level} representations to
learn discursive and stylistic patterns as opposed to simply topics of
discussion, we incorporate a single additional latent vector for the entire thread that interacts with each comment, explaining word choices that are mainly topical rather than discursive or stylistic.
As we demonstrate in our experiments (see Section~\ref{sec:qual})
the \textit{thread-level} embedding learns
distributions more reminiscent of what a traditional
topic model would uncover, while the comment-level embeddings
model styles of speaking and mannerisms that do not directly
indicate specific subjects of conversation.
The joint probability is defined in terms of an energy function that scores latent embeddings and observed word counts across the tree of comments within a thread using log-bilinear potentials, and is globally normalized over all word count and embedding combinations.

\subsection{Probability Model}
\label{sec:prob}

More formally, consider a thread containing $N$ comments each of size $D_{n}$
with a vocabulary of size $K$.
As depicted in Figure \ref{fig:tabm}, each comment is viewed as a bag-of-words, densely connected via a log-bilinear potential to a latent embedding of size $F$.
Let each comment be represented as
as an integer vector $x_n \in \mathbb{Z}^{K}$ where $x_{nk}$ is
number of times word $k$ was observed in comment $n$,
and let $h_{n} = \{0,1\}^{F}$ be the topic embedding for each comment,
and let $h_0 = \{0,1\}^{F}$ be the embedding for the entire thread.
To model topic transitions, we score the embeddings of parent-child pairs with a separate coupling potential as shown in Figure~\ref{fig:tabm} (comments with no parents or children receive additional start/stop biases respectively).
Let replies be represented with sets
$R$, $P_N$, and $C_N$ where $(n,m) \in R$ and $n \in P_m$ and $m \in C_n$ if comment $m$ is a reply to comment $n$.
DDTM assigns probability to a specific configuration of $x,h$ with
an energy function scored by
the emission ($\pi_\text{e}$) and coupling ($\pi_\text{c}$) potentials.
\begin{equation}
\begin{split}
E(x,h;\theta) &= \underbrace{\sum_{n=1}^{N} \pi_\text{e}(h,x,n)}_\text{Emission Potentials}
         + \underbrace{\sum_{(n,m) \in R} \pi_\text{c}( h,n,m)}_\text{Coupling Potentials} \\
\pi_\text{e}(h,x,n) &= h_n^{\intercal} U x_n + x_n^{\intercal} a  + D_n h_n^{\intercal} b \\
&+ h_{0}^{\intercal} V x_n + D_n h_{0}^{\intercal} c \\
\pi_\text{c}(h,n,m) &= h_n^{\intercal} W h_m
\end{split}
\end{equation}
Note that the bias on embeddings is scaled by the number of words in the comment,
which controls for their highly variable length. The joint probability is computed by exponentiating the energy and dividing by a normalizing constant.

\begin{equation}
\begin{split}
p(x,h;\theta) &= \frac{\exp(E(x,h;\theta))}{Z(\theta)} \\
Z(\theta) &= \sum_{x',h'} \exp(E(x',h';\theta))
\end{split}
\end{equation}

This architecture encourages the model to learn
discursive maneuvers via the coupling potentials while
separating within-thread variance and across-thread variance
through the comment-level and thread-level embeddings respectively.
The coupling of latent variables makes factored inference impossible,
meaning that even the exact computation of the partition function
is no longer tractable. This necessitates approximating the gradients
for learning which we will now address.

\section{Learning and Inference}

\begin{figure*}[th!]
\begin{minipage}[c]{0.75\textwidth}
\includegraphics[width=\textwidth]{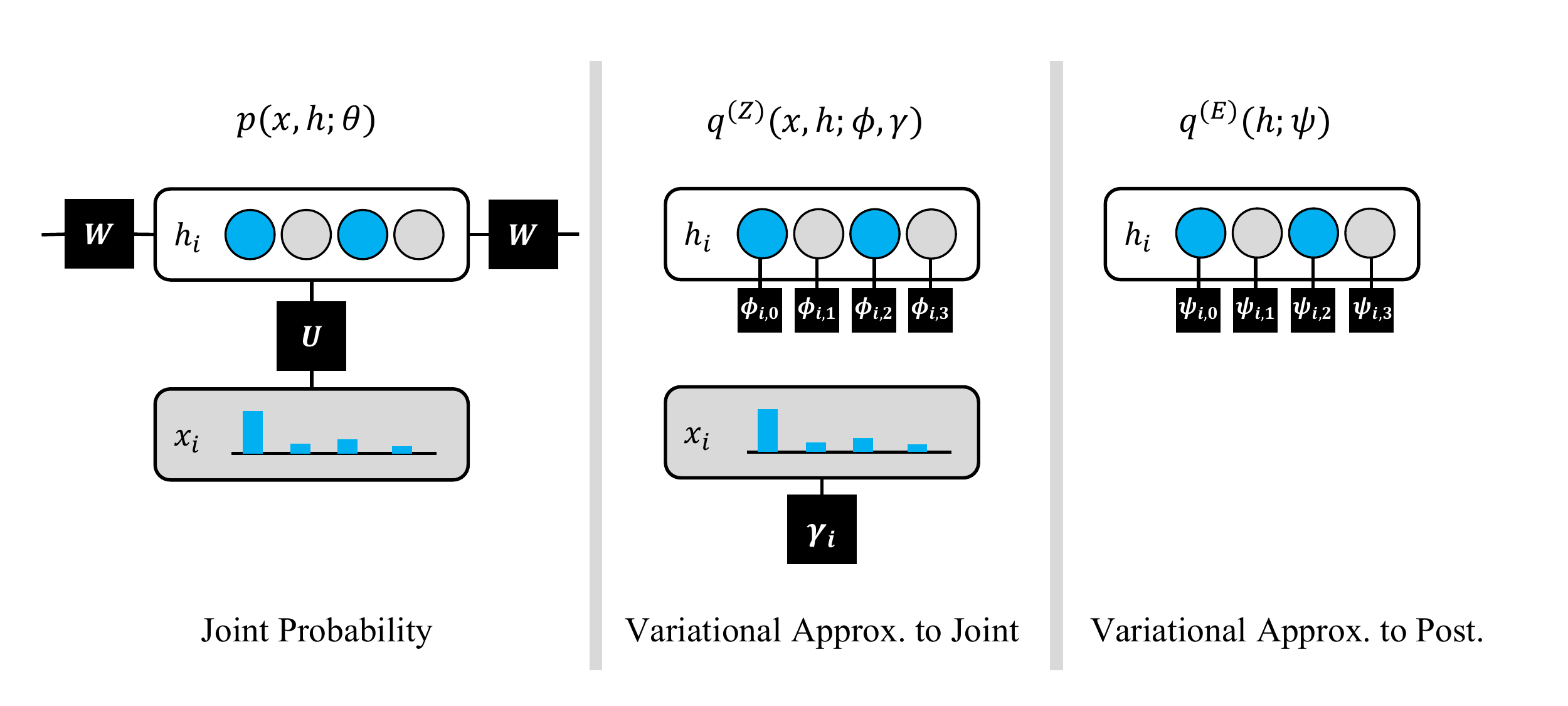}
\end{minipage}
\begin{minipage}[c]{0.2\textwidth}
\caption{Factor graph of full joint compared to mean-field approximations to joint and posterior.}
\label{fig:variational}
\end{minipage}
\end{figure*}

Inference in this model class in intractable, so as has been done
in previous work on topic modeling \citep{LDA} we rely on variational
methods to approximate the gradients needed during training as well as the posteriors
over the topic bit vectors. Specifically, we will need the gradients of the normalizer and the sum of the energy function over the hidden variables
\begin{equation}
E(x;\theta) = \log \sum_{h} \exp ( E(x,h;\theta) )
\end{equation}
which we refer to as the \textit{marginal energy}. Following the approach described for undirected models by~\citet{Jason}, we approximate these quantities and their gradients with respect to the model parameters $\theta$ as we will now describe
(thread-level embeddings are omitted in this section for clarity).

\subsection{Normalizer Approximation}

We aim to train our model to maximize the marginal likelihood of the observed comment word counts, conditioned on the reply links.
To do this we must compute the gradient of the normalizer $Z(\theta)$.
However, this quantity is computationally intractable, as it contains a summation over
all exponential choices for every word in the thread.
Therefore, we must approximate $Z(\theta)$.
Observe that under Jensen's Inequality, we can form the following lower
bound on the normalizer using an approximate joint distribution $q^{(\mathrm{Z})}$.

\begin{equation}
\begin{split}
&\log Z(\theta) = \log \sum_{x,h} \exp(E(x,h;\theta)) \\
                 &\geq \mathbb{E}_{q^{(\mathrm{Z})}} [E(x,h;\theta)]
                 - \mathbb{E}_{q^{(\mathrm{Z})}} [\log q^{(\mathrm{Z})}(x,h;\phi,\gamma)]
\end{split}
\end{equation}

We now define $q^{(\mathrm{Z})}$ as depicted in Figure~\ref{fig:variational} as a mean-field approximation that treats all variables as independent.
We parameterize $q^{(\mathrm{Z})}$
with $\phi_{nf} \in [0,1]$, independent Bernoulli parameters representing the
probability of $h_{nf}$ being equal to $1$, and $\gamma_{nk}$ replicated softmaxes
representing the probability of a word in $x_{n}$ taking the value $k$.
Note that all words in $x_{n}$ are modeled as samples from this single distribution.
The approximation then factors as follows:

\begin{equation}
\begin{split}
&q^{(\mathrm{Z})}(x,h;\phi,\gamma) = q^{(\mathrm{Z})}(x;\gamma) \cdot q^{(\mathrm{Z})}(h;\phi) \\
&q^{(\mathrm{Z})}(x;\gamma) = \prod_{n=1}^{N} \prod_{k=1}^{K} (\gamma_{nk})^{x_{nk}} \\
&q^{(\mathrm{Z})}(h;\phi) = \prod_{n=1}^{N} \prod_{f=1}^{F} \left( (\phi_{nf})^{h_{nf}} (1 - \phi_{nf})^{(1 - h_{nf})} \right)
\end{split}
\end{equation}

We optimize the parameters of $q^{(\mathrm{Z})}$
to maximize its variational lower bound,
via iterative mean-field updates, which allow us to perform
coordinate ascent over the parameters of $q^{(\mathrm{Z})}$.
Maximizing the lower bound with respect to particular $\phi_{nf}$ and $\gamma_{nk}$ while holding all other
parameters frozen, yields the following mean-field update equations (biases omitted for clarity):

\begin{equation}
\begin{split}
\phi_{n\cdot} &= \sigma \left( U \gamma_{n} + \sum_{m \in C_n} W \phi_{m}
                     + \phi_{P_n}^{\intercal} W \right) \\
\gamma_{n\cdot} &= \sigma \left( \phi_{n}^{\intercal} U \right)
\end{split}
\end{equation}

We iterate over the parameters of $q^{(\mathrm{Z})}$ in an ``upward-downward'' manner;
first updating $\phi$ for all comments
with no children, then all comments whose children have been updated,
and so on up to the root of the thread.
Then we perform the same updates in reverse order. After updating all $\phi$, we then update $\gamma$
simultaneously (the components of $\gamma$ are independent conditioned on $\phi$).
We iterate these upward-downward passes until convergence.

\subsection{Marginal Energy Approximation}

We can now approximate the normalizer, but
still need the marginal data likelihood in order to take gradient steps on it and train our model.
In order to recover the marginal likelihood, we must next approximate the
marginal energy $E(x;\theta)$
as it too is intractable. This is due to the coupling potentials, which make the topics across comments dependent even when conditioned on the
word counts. To do this, we form an additional variational approximation (see Figure~\ref{fig:variational})
to the marginal energy, which we optimize similarly.

\begin{equation}
\begin{split}
&E(x;\theta) = \log \sum_{h} \exp(E(x,h;\theta)) \\
              &\geq \mathbb{E}_{q^{(\mathrm{E})}} [E(x,h;\theta)] - \mathbb{E}_{q^{(\mathrm{E})}} [\log q^{(\mathrm{E})}(h;\psi)]
\end{split}
\end{equation}

Since $q^{(\mathrm{E})}(h;\psi)$ need only model the hidden units $h$, we can parameterize
it in the same manner
as $q^{(\mathrm{Z})}(h;\phi)$. Note that while these distributions factor similarly,
they do not share parameters, although we find that in practice, initializing
$\phi \leftarrow \psi$
improves our approximation. We optimize the lower bound on
$E(x;\theta)$ via a similar coordinate ascent strategy, where the mean-field updates take the following form
(biases omitted for clarity):

\begin{equation}
\begin{split}
\psi_{n\cdot} = \sigma \left( U h_{n} + \sum_{m \in C_n} W \psi_{m}
                     + \psi_{P_n}^{\intercal} W \right)
\end{split}
\end{equation}

We can use $q^{(\mathrm{E})}$ to perform inference at test time in our model,
as its parameters $\psi$ directly correspond to the expected values
of the hidden topic embeddings under our approximation.

\subsection{Learning via Gradient Ascent}

We train the parameters of our true model $p(x,h;\theta)$ via stochastic updates
wherein we optimize both approximations on a single datum (i.e. thread) to compute
the approximate gradient of its log-likelihood, and take a single
gradient step on the model parameters (repeating on all training instances
until convergence).
That gradient is given by the difference in feature
expectations under the approximations (entropy terms from the lower
bounds are dropped as they do not depend on $\theta$).

\begin{equation}
\begin{split}
\nabla \log p(x;\theta) &\approx \mathbb{E}_{q^{(\mathrm{E})}(h;\psi)} \left[ \nabla E(x,h;\theta) \right] \\
						&- \mathbb{E}_{q^{(\mathrm{Z})}(x',h;\psi)} \left[ \nabla E(x',h;\theta) \right]
\end{split}
\end{equation}

In summary, we use two separate mean-field approximations to compute lower bounds
on the marginal energy $E(x,h;\theta)$, and its normalizer $Z(\theta)$,
which lets us approximate the marginal likelihood $p(x;\theta)$. Note that as
our estimate on the marginal likelihood
is the difference between two lower bounds, it is not a lower bound itself, although
in practice it works well for training.

\subsection{Scalability and GPU Implementation}

Given the magnitude of our dataset,
it is essential to be able to train
efficiently at scale. Many commonly used topic models such
as LDA~\cite{LDA} have difficulty scaling,
particularly if trained via MCMC methods.
Improvements have been shown from online training~\cite{VBLDA},
but extending such techniques to model comment-to-comment
connections and leverage GPU compute is nontrivial.

In contrast, our proposed model and mean-field procedure can
be scaled efficiently to large data because they are amenable to GPU implementation.
Specifically, the described inference procedure
can be viewed as the output
of a neural network. This is because DDTM is globally normalized
with edges parameterized as log-bilinear weights, which
results in the mean-field updates taking the form of matrix operations
followed by nonlinearities. Therefore, a single iteration of mean-field
is equivalent to a forward pass through a recursive neural network, whose architecture
is defined by the tree structure of the thread. Multiple iterations
are equivalent to feeding the output of the
network back into itself in a recurrent manner,
and optimizing for $T$ iterations is achieved by unrolling the network over $T$ timesteps.
This property makes DDTM highly amenable to efficient training on a GPU, and allowed
us to scale experiments to a dataset of over $13$M total Reddit comments.

\section{Experimental Setup}

\subsection{Data}

We mined a corpus of Reddit threads pulled through the platform's API.
Focusing on the twenty most popular subreddits (gifs, todayilearned, CFB, funny, aww, AskReddit, BlackPeopleTwitter, videos, pics, politics, The\_Donald, soccer, leagueoflegends, nba, nfl, worldnews, movies, mildlyinteresting, news, gaming) over a one month period
yielded $200,000$ threads consisting of $13,276,455$ comments total.
The data was preprocessed by removing special characters, replacing URLs with a domain-specific token,
stemming English words using a Snowball English Stemmer~\cite{Snowball}, removing stopwords,
and truncating the vocabulary to only include the top $10,000$ most common words.
OPs are modeled as a comment at the root of each thread to which all top-level comments respond.
This dataset will be made available for public use after publication.

\subsection{Baselines and Comparisons}

We compare to baselines of Replicated Softmax (RS)~\cite{RS}
and Latent Dirichlet Allocation (LDA)~\cite{LDA}.
RS is a distributed topic model similar to our own, albeit
without any coupling potentials. LDA is a locally normalized
topic model which defines topics as non-overlapping distributions
over words.
To ensure that DDTM does not gain an unfair advantage
purely by having a larger embedding space,
we divide the dimensions equally between comment-
and thread-level.
Unless specified $64$ bits/topics were used.
We experiment with RS and LDA
treating either comments or full threads as documents.

\subsection{Training and Initialization}

SGD was performed using the Adam optimizer~\cite{Adam}. When running inference, we found
convergence was reached in an average of $2$ iterations of updates.
Using a single NVIDIA Titan X (Pascal) card, we were able to train our model to
convergence on the training set of $10$M comments in less than $30$ hours.
It is worth noting that we found DDTM to be fairly sensitive to initialization. We
found best results from Gaussian noise, with comment-level emissions at variance
of $0.01$, thread-level emissions at $0.0001$, and transitions at $0$.
We initialized all biases to $0$ except for the bias on word counts, which we set to
the unigram log-probabilities from the train set.

\section{Results}

\begin{table}[t]
\begin{tabular*}{0.45\textwidth}{@{}lcccc@{}}
\toprule
& \multicolumn{4}{c}{\textbf{Perplexity} (nats)} \\
Bits  & 32 & 64 & 96 & 128 \\
\midrule
RS (thr)    & 2240 & 2234 & 2233 & 2257 \\
RS (cmt)    & 1675 & 1894 & 2245 & 2518 \\
\midrule
DDTM (-cpl) & 2027 & 1704 & 1766 & 1953 \\
DDTM & \textbf{1624} & \textbf{1590} & \textbf{1719} & \textbf{713} \\
\bottomrule
\end{tabular*}
\caption{Perplexity of DDTM with and without coupling potentials (-cpl)
		 vs. baselines trained
         at comment (cmt) or thread (thr) level across various numbers of
         topics and bits. For reference, a unigram model achieves $2644$.}
\label{tab:perp}
\end{table}

\subsection{Evaluating Perplexity}

We compare models by perplexity on a held-out test set, a standard evaluation for generative and latent variables models.

\noindent{\textbf{Setup:}} Due to the use of mean-field approximations for both the marginal energy and normalizer
we lose any guarantees regarding the accuracy of our
likelihood estimate (both approximations are lower bounds, and therefore their difference
is neither a strict lower bound nor guaranteed to be unbiased).
To evaluate perplexity in a more principled way, we use Annealed Importance Sampling (AIS)
to estimate the ratio between our model's normalizer
and the tractable normalizer of a base
model from which we can draw true independent samples
as described by~\citet{AIS}. Note that since the marginal energy is intractable in our
model, unlike a standard RBM, we must sample the joint - and not the marginal
- intermediate distributions.
This yields an unbiased estimate of the normalizer.
The marginal energy must still be approximated
via a lower bound, but given that AIS is unbiased and empirically
low in variance, we can treat
the overall estimate as a lower bound on likelihood for evaluation.
Using $2000$ intermediate distributions, and averaging over $20$ runs, we evaluated
per-word perplexity over a set of $50$ unseen threads. Results are shown in Table~\ref{tab:perp}.

\noindent{\textbf{Results:}} DDTM achieves the lowest perplexity at all dimensionalities.
Note our ablation with the coupling potentials removed (-cpl),
increases perplexity noticeably, indicating that modeling replies helps beyond simply
modeling threads and comments jointly, particularly at larger embeddings.
For reference, a unigram model achieves $2644$.
We find that LDA's approximate perplexity is even worse, likely due
to slackness in its lower bound.

\begin{table}[t!]
\begin{tabular*}{0.45\textwidth}{@{}lcccc@{}}
\toprule
Task  & \textbf{Upvote Regr.} & \textbf{Deletion Pred.} \\
       & (MSE) & (\% acc.) \\
\midrule
LDA (thr)   & 1.952     & 68.35 \\
LDA (cmt)   & 2.047     & 59.26 \\
\midrule
RS (thr)    & 2.024     & 69.92 \\
RS (cmt)    & 2.007     & 66.45 \\
\midrule
DDTM & \textbf{1.933} & \textbf{70.39}  \\
\bottomrule
\end{tabular*}
\caption{Performance of DDTM vs.
		 Replicated Softmax (RS) and Latent Dirichlet Allocation (LDA) at
         predicting upvotes and child deletion.}
\label{tab:pred}
\end{table}

\subsection{Upvote Regression}

To measure how well embeddings capture comment-level
characteristics, we feed them into a linear regression model that predicts the number of upvotes the comment received. Upvotes provide a loose human-annotated measure of likability. We expect that context matters in determining how well received a comment is; the same comment posted in response to different parents may receive a very different number of upvotes.
Hence, we expect comment-level embeddings to be more informative for this task when connected
via our model's coupling potentials.

\noindent{\textbf{Setup:}} We trained a standard linear regressor for each model.
The regressor was trained using ordinary least squares 
on the entire training set of comments using the model's computed topic embeddings as input,
and the number of upvotes on the comment
as the output to predict. As a preprocessing step,
we took the log of the absolute number of votes before training.
We compared models by mean squared error (MSE) on our test set.
Results are shown in Table~\ref{tab:pred}.

\noindent{\textbf{Results:}} DDTM achieves lowest MSE. To assess statistical significance, we performed a $500$ sample bootstrap of our training set. The standard errors of these replications are small, and a two-sample t-test rejects the null hypothesis that DDTM has an average MSE equal to that of the next best method ($p<.001$). Note that our model outperforms both comment- and thread-level embeddings, suggesting that modeling these jointly, and modeling the effect of neighboring representations in the comment graph, more accurately learns information relevant to a comment's social impact.

\subsection{Deletion Prediction}

Comments that are excessively
provocative or in violation of
site rules are often deleted, either by the author or a moderator. We can measure whether DDTM
captures discursive interactions that lead to
such intervention by training a logistic classifier
that predicts whether any of a given comment's children
have been deleted.

\noindent{\textbf{Setup:}}
For each model, a logistic regression classifier was trained stochastically with the Adam optimizer
on the entire training set of comments using the model's computed topic embeddings as input,
and a binary label for whether the comment had any deleted children
as the output to predict. We compared models by accuracy on our test set.
Results are shown in Table~\ref{tab:pred}.

\noindent{\textbf{Results:}} DDTM gets the highest accuracy. Interestingly, thread-level models
do better than comment-level ones, which suggests that certain topics or
even subreddits may correlate with comments being deleted. This makes
sense given that subreddits vary in severity of moderation.
DDTM's performance also demonstrates that modeling comment-to-comment interaction
patterns is helpful in predicting when a comment will spawn a deleted future response,
which strongly matches our intuition.

\subsection{Document Retrieval}

\begin{figure}
\centering
\includegraphics[width=0.5\textwidth]{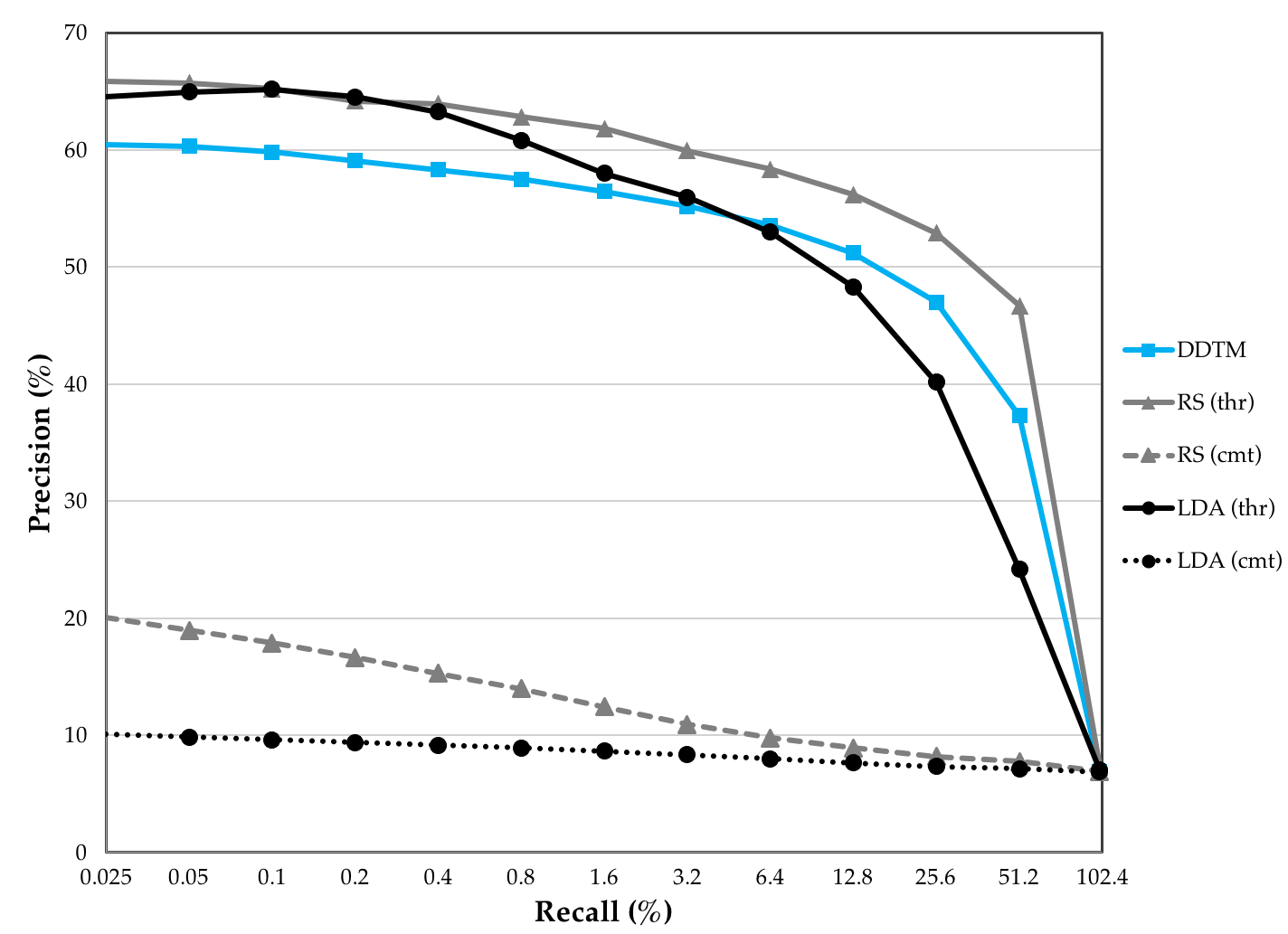}
\caption{Precision vs. recall for document retrieval based on subreddit comparing various models
         for $1000$ randomly selected held-out query comments.}
\label{fig:docret}
\end{figure}

Finally, while DDTM is not designed to better capture topical structure,
we evaluate the extent to which it can still capture this information by performing
document retrieval, a standard evaluation, for which we treat the subreddit
to which a thread was posted as a label for relevance.
Note that every comment within the same thread belongs to the same subreddit,
which gives thread-level models an inherent advantage at this task.
We include this task purely for the purpose of demonstrating that by capturing discursive
patterns, DDTM does not lose the ability to model thread-level topics as well.

\begin{figure}[t!]
\centering
\scalebox{0.6}{\includegraphics[width=0.5\textwidth]{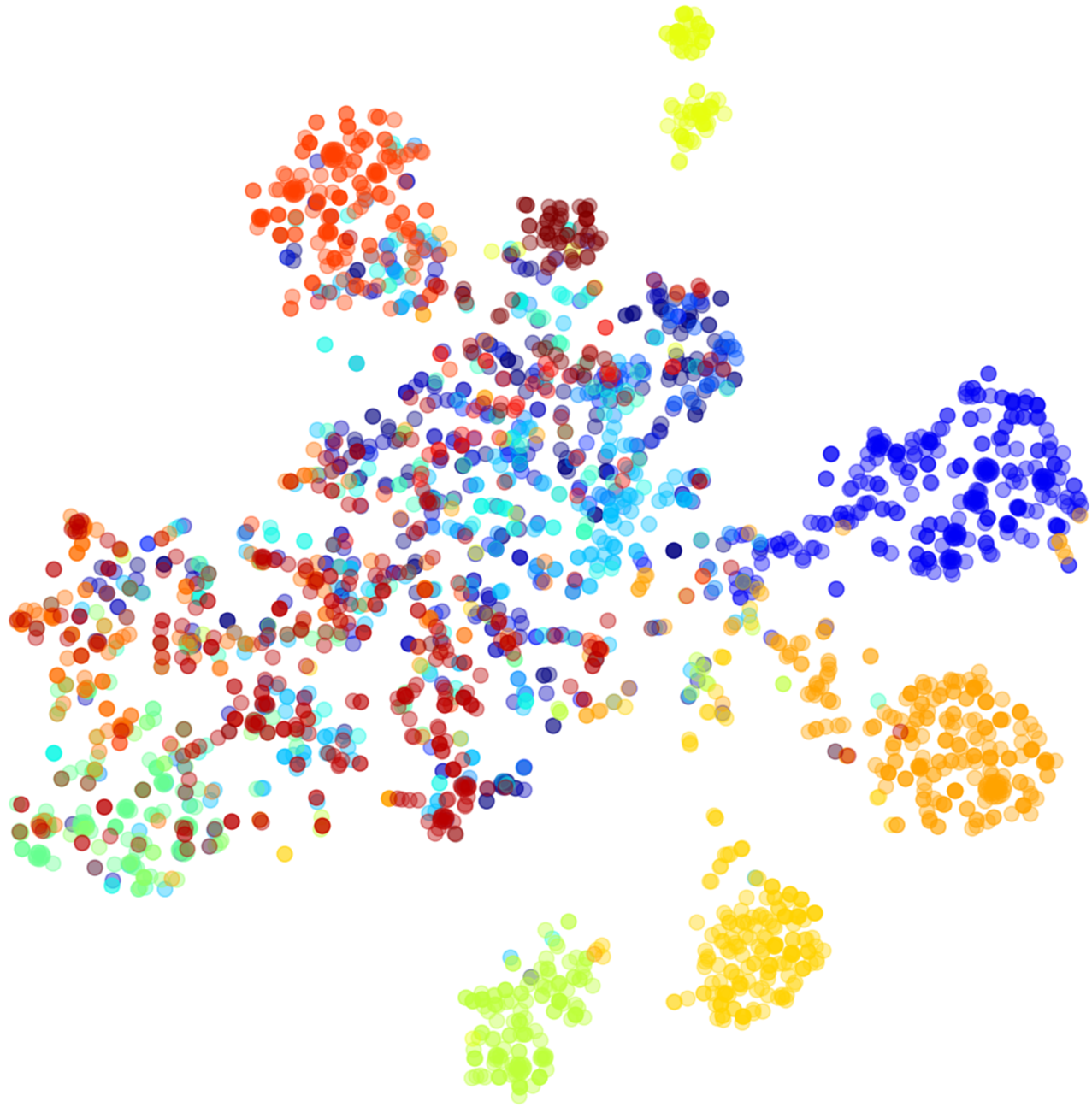}}
\caption{t-SNE visualization of a random sample of DDTM thread-level embeddings
		 colored by subreddit (not observed in training)}
\label{fig:tsne_thread}
\end{figure}

\begin{figure}[t!]
\centering
\scalebox{0.6}{\includegraphics[width=0.5\textwidth]{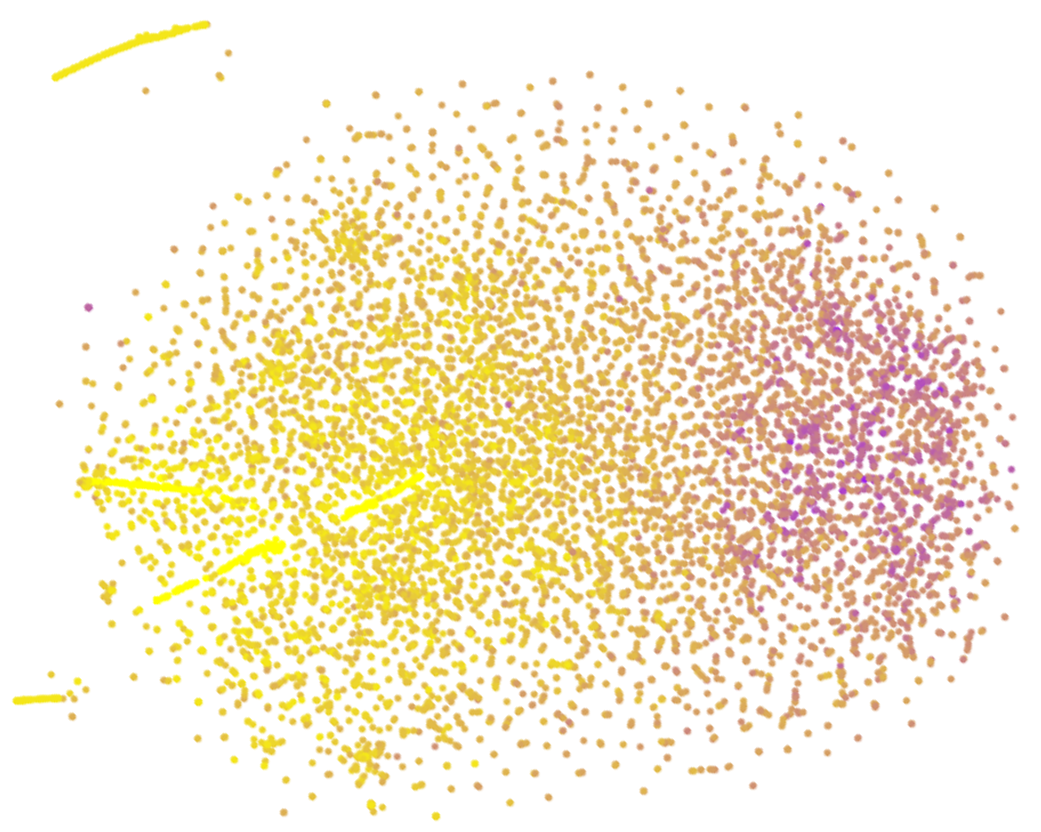}}
\caption{t-SNE visualization of a random sample of DDTM comment-level embeddings
		 colored by log of comment length (darker is longer).}
\label{fig:tsne_comment}
\end{figure}

\noindent{\textbf{Setup:}} Given a query comment from our held-out test set,
we rank the training set by the Dice similarity of the hidden
embeddings computed by the model.
We consider a retrieved comment relevant to the query if they both originate from
the same subreddit, which loosely categorizes the semantic content.
Tuning the number of documents we return allows us to form
precision recall curves, which we show in Figure~\ref{fig:docret}.

\begin{table*}[th!]
\scalebox{0.7}{
\begin{tabular*}{1.3\textwidth}{@{}r|l@{}}
\toprule
\textbf{Bit \#} & \textbf{Associated Word Stems by Emission Weight (Higher Score $\rightarrow$ Lower Score)} \\
\midrule
\multicolumn{2}{l}{Comment-Level}
\vspace{3pt} \\
Bit 1 & \texttt{faq tldrs pms 165 til keyword questions feedback chat pm} \\
2 & \texttt{irl riamverysmart legend omfg riski aboard favr madman skillset tunnel} \\
3 & \texttt{lotta brah ouch spici oof bummer buildup viewership hd uncanni} \\
4 & \texttt{funniest mah tfw teleport fav hoo plz bah whyd dumbest} \\
5 & \texttt{handsom hipster texan hottest whore norwegian shittier scandinavian jealousi douch} \vspace{3pt} \\
\midrule
\multicolumn{2}{l}{Thread-Level}
\vspace{3pt} \\
Bit 1 & \texttt{btc gameplay tutori cyclist dev currenc kitti bitcoin rpg crypto} \\
2 & \texttt{url\_youtu url\_leagueoflegends url\_businessinsider url\_twitter url\_redd url\_snopes} \\
3 & \texttt{comey pede macron pg13 maga globalist ucf committe cuck distributor} \\
4 & \texttt{maduro venezuelan ballot puerto catalonia rican quak skateboard venezuela quebec} \\
5 & \texttt{nra scotus opioid cheney nevada metallica marijuana vermont colorado xanax} \vspace{3pt} \\
\bottomrule
\end{tabular*}}
\caption{Words with the highest emission weight for various comment-level and thread-level bits.}
\label{tab:topics}
\end{table*}

\noindent{\textbf{Results:}} DDTM outperforms both comment-level baselines and is competitive with
thread-level models, even beating LDA at high levels of recall.
This indicates that despite using half of its dimensions to model
comment-to-comment interactions DDTM can still do almost as good a
job of modeling thread-level
semantics as a model using its entire capacity to do so.
The gap between comment-level RS and LDA is also consistent with LDA's known issues
dealing with sparse data~\cite{sridhar2015unsupervised}, and lends credence to our
theory that distributed topic representations are better suited to such domains.

\section{Qualitative Analysis of Topics}

\label{sec:qual}

We now offer qualitative analysis of the topic embeddings learned by our model.
Note that since we use distributed embeddings, our bits
are more akin to filters than complete distributions over words, and we typically
observe as many as half of them active for a single comment.
In a sense, we have an exponential number of topics, whose parameterization
simply factors over the bits.
Therefore, it can
be difficult to interpret them as one would interpret topics learned by a
model such as LDA. Furthermore, we find that in practice this effect is correlated with the topic embedding
size; the more bits our model has, the less sparse and consequently less individually
meaningful the bits become. Therefore for this analysis, we specifically
focus on DDTM trained with 64 bits total.

\subsection{Bits in Isolation}

Directly inspecting the emission parameters, reveals that the comment-level and thread-level
halves of our embeddings capture substantially different aspects of the data
(shown in Table~\ref{tab:topics}) akin to vertical, within-thread, and horizontal, across-thread
sources of variance respectively.
The comment-level topic bits tend to reflect styles of speaking, lingo,
and memes that are not unique to a particular subject of discourse or even subreddit.
For example, comment-level Bit 2 captures many words typical of taunting Reddit comments; replying
with ``/r/iamverysmart'' (a subreddit
dedicated to mocking people who make grandiose claims about their intellect)
is a common way of jokingly implying that
the author of the parent comment takes themselves too seriously
— and thus corresponds to a certain kind of rhetorical move. Further, it is grouped with other words that indicate related rhetorical moves; calling a user ``risky'' or a ``madman'' is
a common means of suggesting that they are engaging in a pointless
act of rebellion.
They also cluster at the coarsest level by length (see Figure~\ref{fig:tsne_comment}) which we
find to correlate with writing style.

By contrast, the thread-level bits are more indicative of specific topics of discussion,
and unsurprisingly they cluster by subreddit (see Figure~\ref{fig:tsne_thread}).
For example, thread-level Bit 3 captures lexicon used almost exclusively
by alt-right Donald Trump supporters as well as the names of various political figures.
Bit 4 highlights words related to civil unrest in Spanish speaking parts of the world.

\subsection{Bits in Combination}

\begin{table*}[th!]
\scalebox{0.7}{
\begin{tabular*}{1.3\textwidth}{@{}r|l@{}}
\toprule
\textbf{Sample \#} & \textbf{Associated Word Stems by Emission Weight (Higher Score $\rightarrow$ Lower Score)} \\
\midrule
\multicolumn{2}{l}{Comment-Level}
\vspace{3pt} \\
Sample 1 & \texttt{grade grader math age 5th 9th 10th till mayb 7th} \\
2 & \texttt{repost damn dope bamboozl shitload imagin cutest sad legendari awhil} \\
3 & \texttt{heh dawg hmm spooki buddi aye m8 aww fam woah} \\
4 & \texttt{hug merci bless tfw prayer pleas dear bear banana satan} \\
5 & \texttt{chuckl cutest funniest yall bummer oooh mustv coolest ok oop} \\
6 & \texttt{cutest heard coolest funniest havent seen ive craziest stupidest weirdest} \\
7 & \texttt{reev keanu christoph murphi walken vincent chris til wick roger} \\
8 & \texttt{moron douchebag stupid dipshit snitch jackass dickhead idioci hypocrit riddanc} \\
9 & \texttt{technic actual realiz happen escal werent citat practic memo cba} \\
10 & \texttt{reddit shill question background user subreddit answer relev discord guild}  \vspace{3pt} \\
\bottomrule
\end{tabular*}}
\caption{Words with the highest emission weight for sample held-out comment reconstructions.}
\label{tab:embeds}
\end{table*}

While these distributions over words (particularly for comment-level bits) can seem
vague, when multiple bits are active, their effects
compound to produce much more specific topics.
One can think of the bits as defining soft filters over the space of words,
that when stacked together carve out patterns not apparent in any of them
individually.
We now analyze a few sample topic embeddings. To do this, we perform
inference as described on a held-out thread, and pass the comment-level
topic embedding for a single sampled comment through our emission matrix and inspect the words with
the highest corresponding weight (shown in Table~\ref{tab:embeds}).
In generative terminology, these can be thought of as reconstructions of comments.

These topic embeddings capture more specific conversational
and rhetorical moves.
For example, Sample 6 displays supportive and interested reactionary language, which one
might expect to see used in response to a post or comment
linking to media or describing something intriguing.
This is of note given that one of the primary aims of including coupling
potentials was to encourage DDTM to learn ``topics'' that correspond to responses
and interactive behavior, something existing methods are largely not designed for.
By contrast, Sample 9 captures a variety of hostile language and insults, which unlike
those discussed previously do not denote membership in a particular
online community. As patterns of toxic and hateful behavior on Reddit are more well-studied
\citep{toxicity},
it could be useful to have a tool to analyze precipitous contexts and parent comments,
something which we hope systems based on coupling of comment embeddings
have the capacity to provide.
Sample 10 is of particular interest as it consists largely of Reddit
terminology. Conversations about the meta of
the site can manifest for example in users accusing each other
of being ``shills'' (i.e. accounts paid to astroturf on behalf of external interests)
or requesting/responding to ``guilding'', a feature which lets users purchase
premium access for each other often in response to a particularly well made comment.

\section{Related Work}

Many topic models such as LDA~\cite{LDA} treat documents as independent mixtures,
yet this approach fails to model how comments
interact with one another throughout a larger discourse if such connections exist in the data.
Other work has considered modeling hierarchy in topics~\cite{griffiths2004hierarchical}. These models form hierarchical representations of topics themselves, but still treat documents as independent. While this approach can succeed in learning topics of various granularities, it does not explicitly track how topics interact in the context of a nested conversation.

Some approaches such as Pairwise-Link-LDA and Link-PSLA-LDA~\cite{nallapati2008joint} attempt to model interactions among documents in an arbitrary graph, albeit with important drawbacks. The former models every possible pairwise link between comments, and the latter models links as a bipartite graph, limiting its ability to scale to large tree-structured threads. Similar work on Topic-Link LDA~\cite{liu2009topic} models link probabilities conditioned on both topic similarity and an authorship model, yet this approach is poorly suited to high volume, semi-anonymous online domains.
Other studies have leveraged reply-structures on Reddit in the context of predicting persuasion~\citep{CMV}, but DDTM
differs in its generative, unsupervised approach.

DDTM's emission potentials are similar to those of Replicated Softmax~\cite{RS}, an undirected model based on a Restricted Boltzmann Machine. Unlike LDA-style models, RS does not assign a topic to each word, but instead builds a distributed representation. In this setting, a single word can be likely under two different topics, both of which are present, and lend probability mass to that word. LDA-style models by contrast would require the topics to compete for the word.

\section{Conclusion}

In this paper we introduce
a novel way to learn topic interactions in
observed discourse trees, and describe GPU-amenable learning techniques
to train on large-scale data mined from Reddit.
We demonstrate improvements
over previous models on perplexity and downstream tasks, and offer
qualitative analysis of learned discursive patterns. The dichotomy
between the two levels of embeddings hints at
applications in style-transfer.

\bibliography{emnlp2018}
\bibliographystyle{acl_natbib_nourl}

\end{document}